\documentclass[letterpaper, 10 pt, journal, twoside]{IEEEtran}

\IEEEoverridecommandlockouts                              

\usepackage{amsmath} 
\usepackage{amssymb}  

\usepackage[dvipdfmx]{graphicx}
\usepackage{url}
\usepackage{multirow}
\usepackage{siunitx}

\newcommand{\figref}[1]{{Fig.~\ref{#1}}}

\newcommand{\bm}[1]{\mbox{\boldmath{$#1$}}}

\title{Whole-body Multi-contact Motion Control \\ for Humanoid Robots Based on \\ Distributed Tactile Sensors}

\author{Masaki Murooka$^{1}$, Kensuke Fukumitsu$^{2}$, Marwan Hamze$^{2}$, Mitsuharu Morisawa$^{1}$, \\Hiroshi Kaminaga$^{1}$, Fumio Kanehiro$^{1}$ and Eiichi Yoshida$^{2}$
  \thanks{Manuscript received: June, 21, 2024; Revised September, 3, 2024, Year; Accepted September, 23, 2024.}
  \thanks{This paper was recommended for publication by Editor Clement Gosselin upon evaluation of the Associate Editor and Reviewers' comments.
    This work was supported by (organizations/grants which supported the work.)}
  \thanks{This work was supported in part by JSPS KAKENHI Grant Number 22H05002.}%
  \thanks{$^{1}$Masaki Murooka, Mitsuharu Morisawa, Hiroshi Kaminaga, and Fumio Kanehiro are with
    CNRS-AIST JRL (Joint Robotics Laboratory), IRL and
    National Institute of Advanced Industrial Science and Technology (AIST),
    1-1-1 Umezono, Tsukuba, Ibaraki 305-8560, Japan.
    {\tt\small \{m-murooka, m.morisawa, hiroshi.kaminaga, f-kanehiro\}@aist.go.jp}}%
  \thanks{$^{2}$Kensuke Fukumitsu, Marwan Hamze, and Eiichi Yoshida are with
    Tokyo University of Science,
    6-3-1 Niijuku, Katsushika-ku, Tokyo 125-8585, Japan.
    {\tt\small 8123539@ed.tus.ac.jp, \{marwan.hamze, eiichi.yoshida\}@rs.tus.ac.jp}}%
  \thanks{Digital Object Identifier (DOI): see top of this page.}%
}

\begin{document}

\maketitle

\setlength{\floatsep}{10pt}
\setlength{\textfloatsep}{12pt}
\setlength{\abovecaptionskip}{4pt}
\setlength{\abovedisplayskip}{6pt}
\setlength{\belowdisplayskip}{6pt}

\begin{abstract}
  To enable humanoid robots to work robustly in confined environments, multi-contact motion that makes contacts not only at extremities, such as hands and feet, but also at intermediate areas of the limbs, such as knees and elbows, is essential.
  We develop a method to realize such whole-body multi-contact motion involving contacts at intermediate areas by a humanoid robot.
  Deformable sheet-shaped distributed tactile sensors are mounted on the surface of the robot's limbs to measure the contact force without significantly changing the robot body shape.
  The multi-contact motion controller developed earlier, which is dedicated to contact at extremities, is extended to handle contact at intermediate areas, and the robot motion is stabilized by feedback control using not only force/torque sensors but also distributed tactile sensors.
  Through verification on dynamics simulations, we show that the developed tactile feedback improves the stability of whole-body multi-contact motion against disturbances and environmental errors.
  Furthermore, the life-sized humanoid RHP Kaleido demonstrates whole-body multi-contact motions, such as stepping forward while supporting the body with forearm contact and balancing in a sitting posture with thigh contacts.
\end{abstract}


\section{Introduction}

\IEEEPARstart{H}{umanoid} robots are expected to realize various manipulation and locomotion tasks to support or replace humans.
To allow humanoid robots to work robustly against disturbances in confined environments, multi-contact motion using whole-body contact is indispensable.
Planning and control of humanoid multi-contact motion has been actively studied in recent years, and various motions such as ladder climbing have been realized~\cite{LadderClimbing:Vaillant:AuRo2016}.
However, in most of the multi-contact motions that have been realized so far, the contact areas of the robot were limited to the hands and feet, and contact was not made with arbitrary areas on the whole body, as is the case with humans.
We refer to such motion with contacts at arbitrary body areas as \textit{whole-body multi-contact motion}.
The challenges for achieving this motion are whole-body contact sensing and balance control.

In this study, we develop a control method to realize whole-body multi-contact motion based on distributed tactile sensors mounted on the robot body surfaces.
Thin and flexible distributed tactile sensors allow measurement of whole-body contact without significantly changing the robot body shape, unlike conventional force/torque sensors.
By extending the previously developed effective multi-contact motion control~\cite{Motion6DoF:Murooka:RAL2022}, we enable the robot to maintain balance while supporting the body with arbitrary areas of the limbs, such as knees and elbows, by equipping the robot with distributed tactile sensors.
Through simulations and real-world experiments, we show that a humanoid robot can perform whole-body multi-contact motion with improved robustness by the developed control method.

\begin{figure*}[tpb]
  \begin{center}
    \includegraphics[width=2.0\columnwidth]{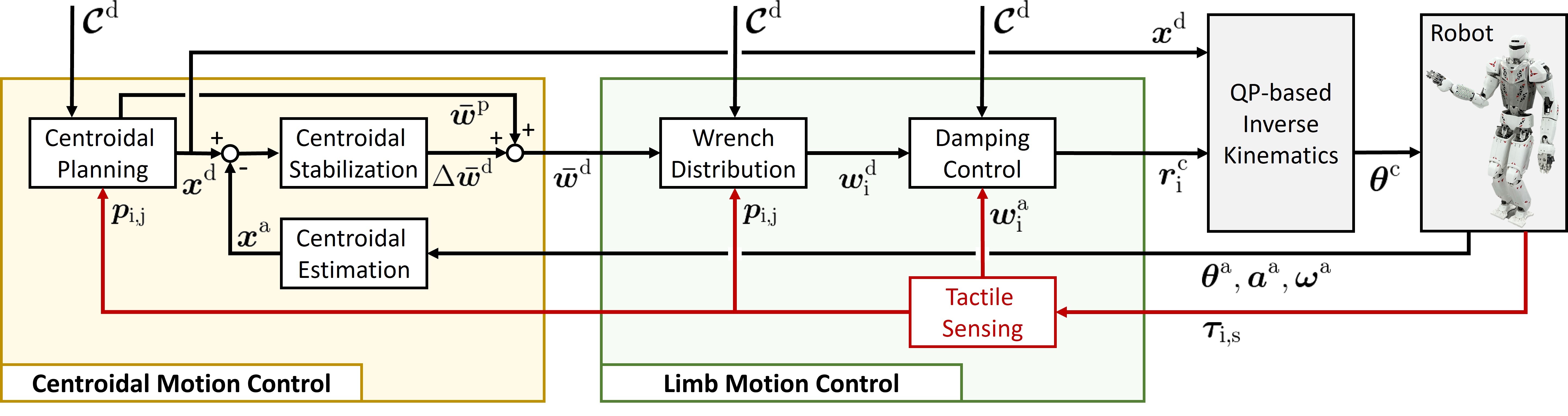}
    \caption{Control system for whole-body multi-contact motion in a humanoid robot.
      \newline
      \footnotesize{
        The control system consists of centroidal motion control and limb motion control.
        Newly added extensions for tactile sensing compared to the previously developed control system~\cite{Motion6DoF:Murooka:RAL2022} are indicated in red in the figure.
        For definitions of the symbols in the figure, see Sections~\ref{sec:centroidal} and \ref{sec:limb}.
      }
    }
    \label{fig:system}
  \end{center}
  \vspace{-4mm}
\end{figure*}

\subsection{Related Works}

\subsubsection{Humanoid Motion Based on Tactile Sensors}

There are a wide range of works on tactile sensing for humanoid robots, from the development of sensors to sensor-based motion generation.
Several tactile sensors have been developed for mounting on the body surface of humanoid robots~\cite{Cheng:RobotSkin:IEEE2019,Maiolino:CapacitiveTactileSystem:IEEESensors2013}, and they have been installed in humanoid robots such as REEM-C~\cite{Cheng:RobotSkin:IEEE2019,REEM-C:PalRobotics} and iCub~\cite{Maiolino:CapacitiveTactileSystem:IEEESensors2013,Natale:iCub:ScienceRobotics2017}.
Tactile sensors mounted on humanoid robots are used to improve the ability of perception and control in various humanoid motions after sensor calibration that includes spatial map construction~\cite{Chefchaouni:TactileSkinCalibration:Sensors2023}.
Typical applications include whole-body interaction with humans~\cite{DeanLeon:TactileComplianceHumanoid:ICRA2019}, whole-body manipulation of large objects~\cite{Mittendorfer:WholebodyTactileManipulation:AdvancedRobotics2015}, in-hand manipulation~\cite{Weiner:TactileGrasp:IROS2021}, and texture classification of the environment~\cite{Taunyazov:TactileTextureIdentification:ICRA2019}.
It should be noted that few previous studies have used tactile sensors for humanoid robot balance control, except for the calculation of the center of pressure (CoP) and support region of the robot's sole in bipedal walking~\cite{Rogelio:TactileWalking:IJHR2020}.
In this study, we explicitly use tactile sensors for balance control in humanoid motion with whole-body contact.

\subsubsection{Multi-contact Motion Control of Humanoid Robots}

Multi-contact motion is an effective strategy for humanoid robots to work stably in complex three-dimensional environments, and planning and control methods have been widely studied~\cite{MultiContactSurvey:Bouyarmane:Reference2018}.
Multi-contact motion by position-controlled humanoid robots often requires compliance control of limbs based on force measurement at the contact areas for stable motion against disturbances and environmental errors.
The contact areas have therefore been inevitably limited to the hands and feet equipped with force/torque sensors~\cite{LadderClimbing:Vaillant:AuRo2016,Motion6DoF:Murooka:RAL2022}.
There are some exceptions where humanoid robots make contact with areas not equipped with force/torque sensors by assuming that the robot motion is quasi-static.
The contact forces are estimated based on the robot's mass model, but in principle, this is limited to slow motions~\cite{Farnioli:WholebodyLocomanipWalkman:IROS2016,MultiContact:Hiraoka:AR2021}.
For torque-controlled humanoid robots, balance control in a posture with knees and elbows in contact with environments has been proposed based on passivity~\cite{MultiContactKnee:Henze:IROS2017}.
Contact at the knees and elbows other than extremities enables activity in confined environments where contact at those areas is unavoidable and improves stability against disturbances by intentionally making wide-area contact.
In this study, we develop a method to broadly realize such whole-body multi-contact motion by position-controlled humanoid robots.

\subsection{Contributions of this Paper}

The contributions of our work are twofold:
(i) we developed a control method for robust multi-contact motion with whole-body contact using distributed tactile sensors mounted on the robot body surface
and (ii) we performed simulation and real-world experiments in which a humanoid robot with limb-mounted distributed tactile sensors demonstrated various whole-body multi-contact motions robustly.
To the best of the authors' knowledge, this is the first study in which a position-controlled life-sized humanoid robot has realized dynamic motion with contact at an intermediate area, such as the forearm, through real-time tactile sensor feedback.

\section{Method Overview} \label{sec:system}

\figref{fig:system} shows the developed control system for whole-body multi-contact motion by humanoid robots.
We extend our previous framework~\cite{Motion6DoF:Murooka:RAL2022} by enhancing modules for centroidal control and limb control to adapt to whole-body contact based on tactile sensors in Sections~\ref{sec:centroidal} and \ref{sec:limb}, respectively.
Then, in Section~\ref{sec:exp}, we present simulation and real-world experiments.
In our control system, distributed tactile sensors are used to obtain the contact region for centroidal planning and wrench distribution, and to obtain the contact wrench for damping control.

We assume that the robot is position-controlled.
The sequence of timing, location, and area (e.g., feet, knee, and elbows) of contact $\bm{\mathcal{C}}^{\mathrm{d}}$ is assumed to be determined manually or by a global planner~\cite{MultiContactSurvey:Bouyarmane:Reference2018}.

\section{Centroidal Motion Control} \label{sec:centroidal}

\begin{figure}[tpb]
  \begin{center}
    \includegraphics[width=0.75\columnwidth]{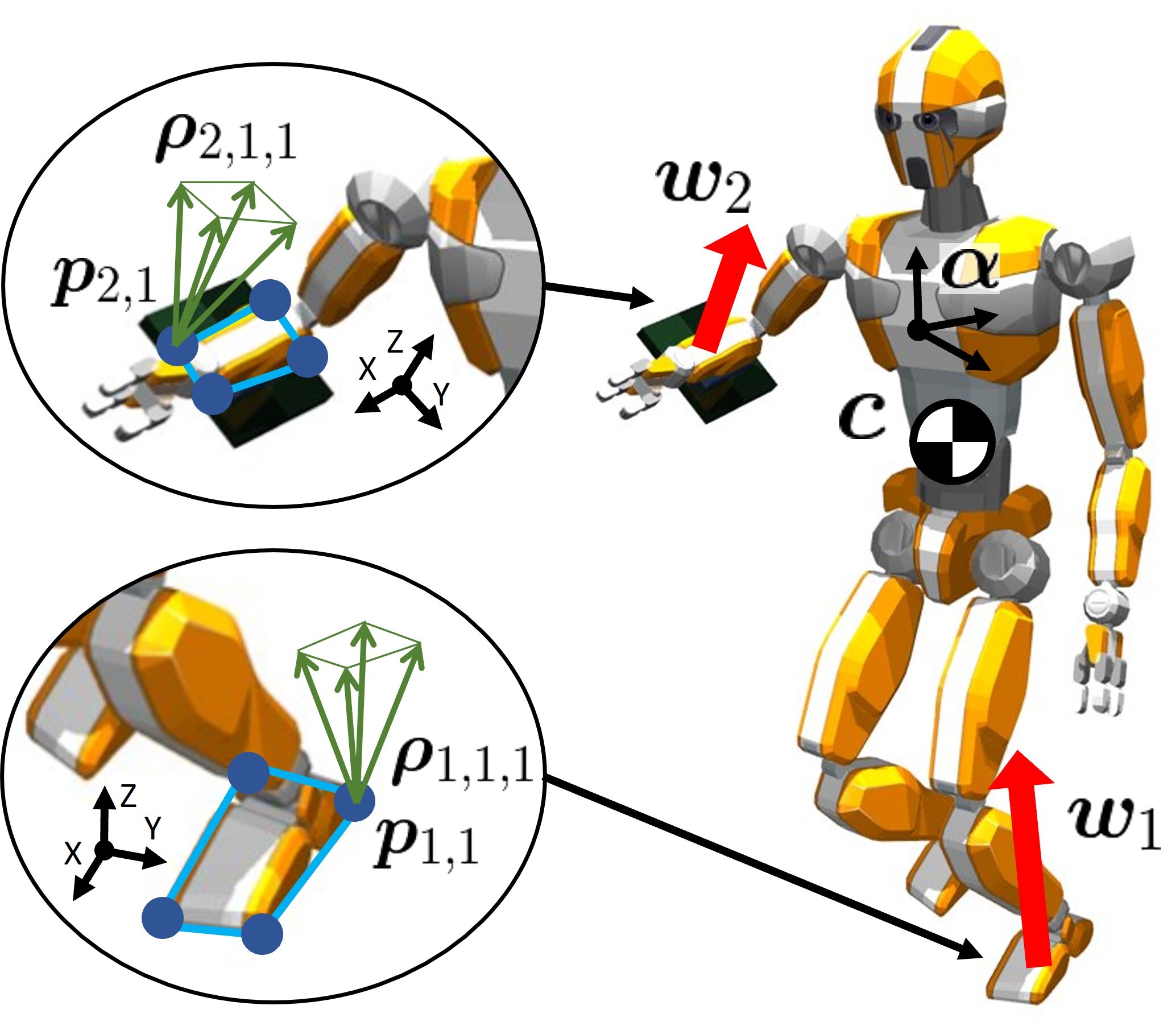}
    \caption{Contact wrench representation.
    }
    \label{fig:resultant-wrench}
  \end{center}
\end{figure}

\subsection{Definition of Resultant Wrench} \label{subsec:resultant-wrench}

The centroidal motion is controlled by the resultant wrench acting on the robot's center of mass (CoM), which is expressed as follows:
\begin{align}
  & \bm{\bar{w}}(\bm{\lambda}, \bm{c}) =
  \sum_i \sum_j \sum_k
  \begin{bmatrix}
    \lambda_{\mathrm{i,j,k}} \bm{\rho}_{\mathrm{i,j,k}} \\
    (\bm{p}_{\mathrm{i,j}} - \bm{c}) \times \lambda_{\mathrm{i,j,k}} \bm{\rho}_{\mathrm{i,j,k}}
  \end{bmatrix} \label{eq:resultant-wrench} \\
  & \mathrm{where} \ \ \bm{\lambda} = \begin{bmatrix} \cdots & \lambda_{\mathrm{i,j,k}} & \cdots \end{bmatrix}^{\mathsf{T}} \nonumber
\end{align}
In this paper, the resultant wrench is represented as $\bm{\bar{w}} \in \mathbb{R}^6$ with a bar added above the symbol.
$\bm{c} \in \mathbb{R}^3$ is the CoM position,
the subscript $\mathrm{i}$ corresponds to each contact area,
$\bm{p}_{\mathrm{i,j}} \in \mathbb{R}^3$ is the $\mathrm{j}$-th vertex of the contact polygon,
$\bm{\rho}_{\mathrm{i,j,k}} \in \mathbb{R}^3$ is the $\mathrm{k}$-th ridge vector of the friction pyramid at the $\mathrm{j}$-th vertex, and
$\lambda_{\mathrm{i,j,k}} \in \mathbb{R}$ is the scale of the contact force along the ridge $\bm{\rho}_{\mathrm{i,j,k}}$.
\figref{fig:resultant-wrench} illustrates these variables.
All variables are represented in world coordinates.
Because $\bm{p}_{\mathrm{i,j}}$ and $\bm{\rho}_{\mathrm{i,j,k}}$ are included in a given target contact sequence, the resultant wrench~\eqref{eq:resultant-wrench} can be regarded as a function of $\bm{\lambda}$ and $\bm{c}$.

Contact with the body area equipped with distributed tactile sensors allows measuring the actual contact polygon between the robot and the environment.
Our control system enables online updating of the contact polygon vertices $\bm{p}_{\mathrm{i,j}}$ when there is a significant difference between the predefined and the measured contact polygons.
By updating contact polygons, it is possible to avoid situations where contact instability occurs due to attempts to generate excessive moments in response to overestimated contact polygons, and situations where sufficient moments to restore balance cannot be exerted in feedback control due to underestimated contact polygons.

\subsection{MPC for Centroidal Planning} \label{subsec:mpc}

The centroidal motion of a humanoid robot is governed by the Newton-Euler equation:
\begin{align}
  \begin{bmatrix}
    m (\bm{\ddot{c}} + \bm{g}) \\
    \bm{I} \bm{\dot{\omega}} + \bm{\omega} \times \bm{I} \bm{\omega}
  \end{bmatrix}
  = \bm{\bar{w}}(\bm{\lambda}, \bm{c})
  \label{eq:newton-euler}
\end{align}
where $m \in \mathbb{R}$ and $\bm{I} \in \mathbb{R}^{3 \times 3}$ are the robot mass and inertia matrix, respectively, and
$\bm{\omega} \in \mathbb{R}^3$ is the angular velocity of the base link.

Equation~\eqref{eq:newton-euler} can be expressed as a nonlinear discrete system with $\bm{x}$ and $\bm{\lambda}$ as the centroidal state and control input, respectively:
\begin{align}
  & \bm{x}[l+1] = \bm{f} \left( \bm{x}[l], \bm{\lambda}[l] \right) \label{eq:state-eq} \\
  & \mathrm{where} \ \ \bm{x} = \begin{bmatrix} \bm{c}^{\mathsf{T}} & \bm{\alpha}^{\mathsf{T}} & \bm{v}^{\mathsf{T}} & \bm{\omega}^{\mathsf{T}} \end{bmatrix}^{\mathsf{T}} \nonumber
\end{align}
where $l \in \mathbb{Z}$ is the index of the control step,
$\bm{\alpha} \in \mathbb{R}^3$ is the Euler angle representing the orientation of the base link, and
$\bm{v} \in \mathbb{R}^3$ is the CoM velocity.
In our control system, the Euler method is used for discretization.

To control the system~\eqref{eq:state-eq}, the optimal control input is calculated to minimize the following objective function in the receding horizon:
\begin{align}
  \min_{\scriptsize \bm{\lambda}} \  \sum_{\tilde{l} = l}^{l + N_h}
  \left( \left\| \bm{x}[\tilde{l}] - \bm{x}^{\mathrm{ref}}[\tilde{l}] \right\|^2 + w_{\mathrm{\lambda}} \left\| \bm{\lambda}[\tilde{l}] \right\|^2 \right)
  \ \ {\rm s.t.} \ \ \bm{\lambda} \geq \bm{0}
  \label{eq:mpc-optimization}
\end{align}
where $N_h \in \mathbb{Z}$ is the number of horizon steps,
$\bm{x}^{\mathrm{ref}} \in \mathbb{R}^{12}$ is the reference of the centroidal state, and
$w_{\mathrm{\lambda}} \in \mathbb{R}$ is the weight of the control input.

By inserting the optimal control input into \eqref{eq:state-eq}, the centroidal state at the next control step $\bm{x}[l+1]$ can be obtained.
This is used as the initial state of model predictive control (MPC) in the next control step as well as the target in the inverse kinematics calculation.

Recently, MPC based on differential dynamic programming (DDP)~\cite{DDP:Tassa:ICRA2014}, which improves computational efficiency through recursion, has been widely used in legged robot motion control~\cite{DDPWalking:Dantec:Humanoids2022,MPCLocoManip:Sleiman:RAL2021}.
Our control system also uses DDP to solve the optimization problem~\eqref{eq:mpc-optimization}.

\subsection{Centroidal Stabilization}

Due to the reduced model used in MPC and external disturbances to the robot, there is always an error between the centroidal state planned by MPC and the actual centroidal state.
To cope with such an error, stabilization control by adjusting the resultant wrench is introduced.

The adjustment amount of the resultant wrench is determined by proportional-derivative (PD) feedback control as follows:
\begin{align}
  \Delta \bm{\bar{w}}^{\mathrm{d}} &=
  \begin{bmatrix}
    \bm{P}_{\mathrm{L}} (\bm{c}^{\mathrm{d}} - \bm{c}^{\mathrm{a}}) + \bm{D}_{\mathrm{L}} (\bm{v}^{\mathrm{d}} - \bm{v}^{\mathrm{a}}) \\
    \bm{P}_{\mathrm{A}} \mathrm{log} \left( \bm{R}^{\mathrm{d}} {\bm{R}^{\mathrm{a}}}^{\mathsf{T}} \right) + \bm{D}_{\mathrm{A}} (\bm{\bm{\omega}}^{\mathrm{d}} - \bm{\bm{\omega}}^{\mathrm{a}})
    \label{eq:centroidal-feedback}
  \end{bmatrix}
\end{align}
where the superscripts $\mathrm{d}$ and $\mathrm{a}$ in the symbols represent the MPC planned value and the sensor measured value, respectively;
$\bm{R}^{\mathrm{d}}$ and $\bm{R}^{\mathrm{a}}$ are the rotation matrices corresponding to $\bm{\alpha}^{\mathrm{d}}$ and $\bm{\alpha}^{\mathrm{a}}$, respectively;
$\mathrm{log}(\bm{R}) \in \mathbb{R}^3$ is a function that converts the rotation matrix $\bm{R}$ to an equivalent axis-angle vector; and
$\bm{P}_{\mathrm{L}}, \bm{D}_{\mathrm{L}}, \bm{P}_{\mathrm{A}}, \bm{D}_{\mathrm{A}} \in \mathbb{R}^{3 \times 3}$ are the diagonal matrices of the feedback gains.

The desired resultant wrench is calculated as follows:
\begin{align}
  \bm{\bar{w}}^{\mathrm{d}} = \bm{\bar{w}}(\bm{\lambda}^{\mathrm{mpc}}, \bm{c}^{\mathrm{d}}) + \Delta \bm{\bar{w}}^{\mathrm{d}}
  \label{eq:desired-resultant-wrench}
\end{align}
where $\bm{\lambda}^{\mathrm{mpc}}$ is the optimal control input of the current control step planned by MPC (i.e., the optimal solution of~\eqref{eq:mpc-optimization}).

\subsection{Centroidal Estimation}

The actual centroidal state in~\eqref{eq:centroidal-feedback} is estimated by an inertial measurement unit (IMU) mounted on the base link\footnote{A detailed description can be found at \url{https://scaron.info/robotics/floating-base-estimation.html}}.
First, an extended Kalman filter estimates the orientation of the base link based on linear acceleration and angular velocity measured by the IMU.
Next, the robot model with the measured joint angles is placed so that the anchor point coincides with the planned position under the estimated orientation of the base link.
The anchor point is a weighted average of the contact points according to the scale of the planned contact force.
Finally, the estimated CoM position is calculated from the robot model.

\section{Limb Motion Control} \label{sec:limb}

\subsection{Wrench Distribution}

The robot distributes the wrench to the contact areas to achieve the desired resultant wrench~\eqref{eq:desired-resultant-wrench} through contacts with the environment.
The wrench distribution is formulated as the following optimization problem:
\begin{align}
  \min_{\scriptsize \bm{\lambda}} \ \left\| \bm{\bar{w}}(\bm{\lambda}, \bm{c}^{\mathrm{d}}) - \bm{\bar{w}}^{\mathrm{d}} \right\|^2 \ \ {\rm s.t.} \ \ \bm{\lambda} \geq \bm{0}
  \label{eq:wrench-distrib}
\end{align}
Because $\bm{\bar{w}}(\bm{\lambda}, \bm{c})$ is linear with respect to $\bm{\lambda}$, the optimization problem~\eqref{eq:wrench-distrib} can be formulated as quadratic programming.

With $\bm{\lambda}^{\mathrm{dist}}$ as the optimal solution in~\eqref{eq:wrench-distrib}, the desired contact wrench at the $\mathrm{i}$-th area is expressed as follows:
\begin{align}
  \bm{w}^{\mathrm{d}}_{\mathrm{i}} =
  \sum_j \sum_k
  \begin{bmatrix}
    \lambda^{\mathrm{dist}}_{\mathrm{i,j,k}} \bm{\rho}_{\mathrm{i,j,k}} \\
    (\bm{p}_{\mathrm{i,j}} - \bm{c}^{\mathrm{d}}) \times \lambda^{\mathrm{dist}}_{\mathrm{i,j,k}} \bm{\rho}_{\mathrm{i,j,k}}
  \end{bmatrix}
  \label{eq:limb-wrench}
\end{align}
Due to the inequality constraint in~\eqref{eq:wrench-distrib}, the distributed contact wrench~\eqref{eq:limb-wrench} satisfies the unilateral and friction constraints.
By updating the contact polygon vertices $\bm{p}_{\mathrm{i,j}}$ based on the measurements of distributed tactile sensors as described in Section~\ref{subsec:resultant-wrench}, it is possible to calculate a desired contact wrench that can be exerted in actual contact, rather than a predefined one.

\subsection{Damping Control}

Damping control~\cite{Stabilizer:Kajita:IROS2010} is applied to achieve the desired contact wrench $\bm{w}^{\mathrm{d}}_{\mathrm{i}}$ of each contact area.

Let $\bm{p}_{\mathrm{i}}^{\mathrm{d}} \in \mathbb{R}^3$ and $\bm{R}_{\mathrm{i}}^{\mathrm{d}} \in \mathbb{R}^{3 \times 3}$ represent the desired pose of the contact area determined from the given target contact sequence, and $\bm{p}_{\mathrm{i}}^{\mathrm{c}}$ and $\bm{R}_{\mathrm{i}}^{\mathrm{c}}$ represent the compliance pose of the contact area.
In damping control, the compliance pose is updated to satisfy the following relationship:
\begin{align}
  & \bm{K}_{\mathrm{d}} \Delta \bm{\dot{r}}_{\mathrm{i}}^{\mathrm{c}} + \bm{K}_{\mathrm{s}} \Delta \bm{r}_{\mathrm{i}}^{\mathrm{c}} = \bm{K}_{\mathrm{f}} (\bm{w}_{\mathrm{i}}^{\mathrm{a}} - \bm{w}_{\mathrm{i}}^{\mathrm{d}}) \label{eq:damping-control} \\
  & \mathrm{where} \ \ \Delta \bm{r}_{\mathrm{i}}^{\mathrm{c}} =
  \begin{bmatrix} \Delta \bm{r}_{\mathrm{i,L}}^{\mathrm{c}} \\ \Delta \bm{r}_{\mathrm{i,A}}^{\mathrm{c}} \end{bmatrix} =
  \begin{bmatrix} \bm{p}_{\mathrm{i}}^{\mathrm{c}} - \bm{p}_{\mathrm{i}}^{\mathrm{d}} \\ \mathrm{log}\left(\bm{R}_{\mathrm{i}}^{\mathrm{c}} {\bm{R}_{\mathrm{i}}^{\mathrm{d}}}^{\mathsf{T}}\right) \end{bmatrix} \nonumber
\end{align}
where $\bm{w}_{\mathrm{i}}^{\mathrm{a}}$ is the measured contact wrench at the $\mathrm{i}$-th contact area, and
$\bm{K}_{\mathrm{d}}, \bm{K}_{\mathrm{s}}, \bm{K}_{\mathrm{f}} \in \mathbb{R}^{6 \times 6}$ are diagonal matrices representing the damping parameter, spring parameter, and wrench gain, respectively.

For discrete-time control, \eqref{eq:damping-control} is implemented as follows:
\begin{align}
  & \Delta \bm{r}_{\mathrm{i,L}}^{\mathrm{c}}[l+1] = \Delta \bm{r}_{\mathrm{i,L}}^{\mathrm{c}}[l] + \Delta t \Delta \bm{\dot{r}}_{\mathrm{i,L}}^{\mathrm{c}}[l] \label{eq:damping-control-imp} \\
  & \Delta \bm{r}_{\mathrm{i,A}}^{\mathrm{c}}[l+1] = \mathrm{log}\left(\mathrm{exp} \left(\Delta t \Delta \bm{\dot{r}}_{\mathrm{i,A}}^{\mathrm{c}}[l] \times \right) \, \mathrm{exp} \left(\Delta \bm{r}_{\mathrm{i,A}}^{\mathrm{c}}[l] \times \right) \right) \nonumber \\
  & \mathrm{where} \ \ \Delta \bm{\dot{r}}_{\mathrm{i}}^{\mathrm{c}}[l] = - \frac{\bm{K}_{\mathrm{s}}}{\bm{K}_{\mathrm{d}}} \Delta \bm{r}_{\mathrm{i}}^{\mathrm{c}}[l] + \frac{\bm{K}_{\mathrm{f}}}{{\bm{K}_{\mathrm{d}}}} (\bm{w}_{\mathrm{i}}^{\mathrm{a}}[l] - \bm{w}_{\mathrm{i}}^{\mathrm{d}}[l]) \nonumber
\end{align}
where $\mathrm{exp}(\bm{a}\times) \in \mathbb{R}^{3 \times 3}$ is a function that converts the axis-angle vector $\bm{a}$ to an equivalent rotation matrix.
Because $\bm{K}_{\mathrm{d}}, \bm{K}_{\mathrm{s}},$ and $\bm{K}_{\mathrm{f}}$ are diagonal matrices, their division implies element-wise computation.

\subsection{Distributed Tactile Sensing}

The actual wrench at the contact area must be measured for damping control.
Although 6-axis force/torque sensors are typically installed in the robot's feet, it is rare to find them in the knees and elbows due to geometrical restrictions.
To allow for contact at such areas, in this study, thin and flexible distributed tactile sensors are used for damping control.

\begin{figure}[tpb]
  \begin{center}
    \includegraphics[width=1.0\columnwidth]{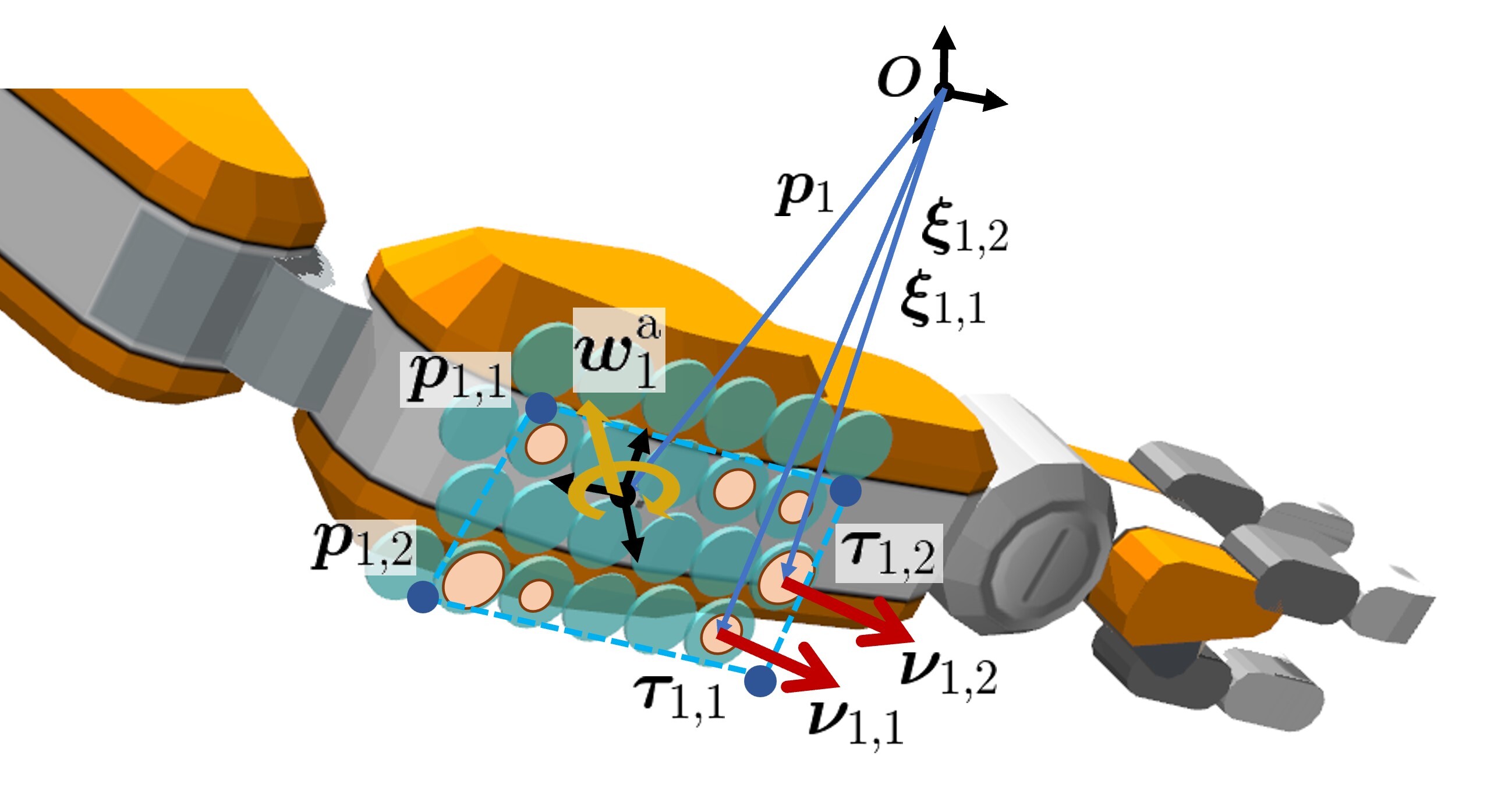}
    \caption{Conversion of distributed tactile intensity to contact wrench.
    }
    \label{fig:tactile-sensing}
  \end{center}
\end{figure}

We assume that the distributed tactile sensors consist of multiple cells that can measure only the tactile response in the normal direction, as shown in~\figref{fig:tactile-sensing}.
Data from tactile sensors can be converted to contact wrench as follows:
\begin{align}
  \bm{w}_{\mathrm{i}}^{\mathrm{a}} =
  \sum_s
  \begin{bmatrix}
    f_{\tau}(\tau_{\mathrm{i,s}}) \, \bm{\nu}_{\mathrm{i,s}} \\
    (\bm{\xi}_{\mathrm{i,s}} - \bm{p}_{\mathrm{i}}) \times f_{\tau}(\tau_{\mathrm{i,s}}) \, \bm{\nu}_{\mathrm{i,s}}
  \end{bmatrix}
  \label{eq:tactile-wrench}
\end{align}
where the subscript $\mathrm{s}$ corresponds to each cell of the distributed tactile sensors;
$\bm{\xi}_{\mathrm{i,s}}$, $\bm{\nu}_{\mathrm{i,s}} \in \mathbb{R}^3$, and $\tau_{\mathrm{i,s}} \in \mathbb{R}$ are the position, normal vector, and measured tactile intensity of the $\mathrm{s}$-th sensor cell at the $\mathrm{i}$-th contact area, respectively; and
$f_{\tau}$ is a sensor-specific function that converts raw tactile intensity to force.
When the cells are arranged on a plane, the force perpendicular to the normal and the moment around the normal are always zero in~\eqref{eq:tactile-wrench}.
For those axes, the damping control parameters are adjusted to disable the wrench feedback.

To update the contact polygon vertices online, the contact surface region is estimated from the measurements of the distributed tactile sensors.
In this study, we assume a contact surface with an axis-aligned rectangular region and calculate the smallest rectangle encompassing all cells where contact is detected.
The calculated rectangle vertices are assigned to $\bm{p}_{\mathrm{i,j}}$ in~\eqref{eq:resultant-wrench} when updating the contact polygon vertices according to the actual contact.

\section{Experiments} \label{sec:exp}

\subsection{Implementation}

\subsubsection{Humanoid Robot with Tactile Sensors}

\begin{figure}[tpb]
  \begin{center}
    \includegraphics[width=1.0\columnwidth]{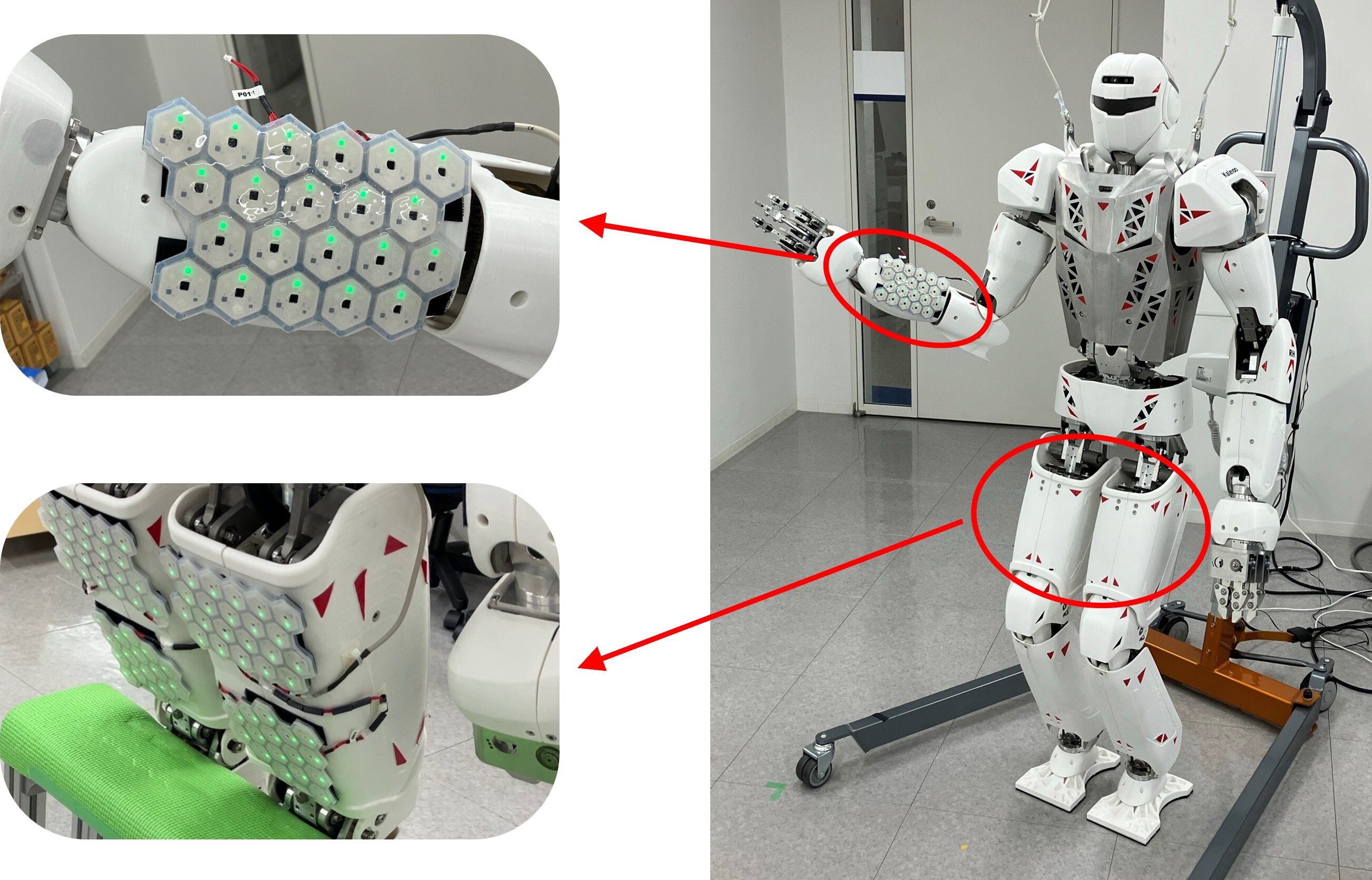}
    \caption{RHP Kaleido with distributed tactile sensors mounted on one forearm and both thighs.
    }
    \label{fig:rhp7-eskin}
  \end{center}
\end{figure}

We mounted distributed tactile sensors e-skin~\cite{Cheng:RobotSkin:IEEE2019} from intouch-robotics on one forearm and both thighs of RHP Kaleido, a life-sized humanoid robot developed by Kawasaki Heavy Industry~\cite{Kakiuchi:RHP:IROS2017}, as shown in~\figref{fig:rhp7-eskin}.
E-skin is a deformable sheet of sensors about 5~mm thick that can be easily mounted on the surface of robot links.
The e-skin patches on the robot consist of 21 and 75 hexagonal-shaped cells in the forearm and both thighs, respectively.
Each cell is equipped with sensors that can measure tactile intensity in the normal direction, proximity, and acceleration.
The relative positions of tactile sensor cells can be automatically calibrated by measuring the gravitational acceleration with different orientations using accelerometers at each cell~\cite{Cheng:RobotSkin:IEEE2019}.
Then the tactile sensor patches were interactively aligned with the robot body based on 3D visualization of the robot model.
The function~$f_{\tau}$ in~\eqref{eq:tactile-wrench}, which converts tactile intensity to force, was identified by linear fitting of pairs of raw e-skin measurements and calibrated force gauge measurements.

\renewcommand{\arraystretch}{1.25}
\begin{table*}[h]
  \caption{Parameters for centroidal MPC \eqref{eq:mpc-optimization} and stabilization \eqref{eq:centroidal-feedback}}
  \label{tab:param-mpc-stabilization}
  \vspace{-4mm}
  \begin{center}
    \begin{tabular}{ccc|cccc}
      \hline
      $w_{\mathrm{\lambda}}$ & $N_h$ & $\Delta \tau$ & $\bm{P}_{\mathrm{L}}$ & $\bm{D}_{\mathrm{L}}$ & $\bm{P}_{\mathrm{A}}$ & $\bm{D}_{\mathrm{A}}$\\
      \hline
      $\num{5e-6}$  & \hspace{2mm}$40$\hspace{2mm} & \hspace{2mm}$0.05$\hspace{4mm} & \hspace{4mm}$\mathrm{diag}(750, 750, 10000)$ & $\mathrm{diag}(150, 150, 150)$ & $\mathrm{diag}(750, 750, 750)$ & $\mathrm{diag}(150, 150, 150)$\\
      \hline
    \end{tabular}\\
    \vspace{2mm}
    \begin{minipage}{1.7\columnwidth}
      \footnotesize{
        $\Delta \tau$ is the discretization period of the MPC horizon, and
        $\mathrm{diag}$ denotes the diagonal matrix.
      }
    \end{minipage}
  \end{center}
  \vspace{-2mm}
  \caption{Parameters for damping control \eqref{eq:damping-control}}
  \label{tab:damping-param}
  \vspace{-4mm}
  \begin{center}
    \begin{tabular}{l||ccc}
      \hline
      & $\bm{K}_{\mathrm{d}}$ & $\bm{K}_{\mathrm{s}}$ & $\bm{K}_{\mathrm{f}}$ \\
      \hline
      Contact phase \hspace{2mm}&\hspace{2mm} $\mathrm{diag}(10000, 10000, 10000, 100, 100, 100)$ & $\mathrm{diag}(0, 0, 0, 0, 0, 2000)$ & $\mathrm{diag}(1, 1, 1, 1, 1, 0)$ \\
      \hline
      Non-contact phase \hspace{2mm}&\hspace{2mm} $\mathrm{diag}(300, 300, 300, 40, 40, 40)$ & $\mathrm{diag}(2250, 2250, 2250, 400, 400, 400)$ & $\mathrm{diag}(0, 0, 0, 0, 0, 0)$ \\
      \hline
    \end{tabular}\\
    \vspace{2mm}
    \begin{minipage}{1.8\columnwidth}
      \footnotesize{
        In the non-contact phase, the compliance displacement~$\Delta \bm{r}_{\mathrm{i}}^{\mathrm{c}}$ converges to zero when we set $\bm{K}_{\mathrm{s}}$ to positive values and $\bm{K}_{\mathrm{f}}$ to zero, as shown in \eqref{eq:damping-control-imp}.
        $\bm{K}_{\mathrm{d}}$ values in the contact phase at the intermediate contact areas of the limb were tuned for each motion in the range of 1000 to 100000 for translation and 100 to 1000 for rotation.
      }
    \end{minipage}
  \end{center}
  \vspace{-4mm}
\end{table*}

\subsubsection{Software Framework}

The control system was implemented in {\tt C++} within a real-time robot control framework {\tt mc\_rtc}~\cite{mc_rtc:github2023}.
In this system, kinematics commands, such as the CoM position, the base link orientation, and the poses of contact areas, are passed to the acceleration-based whole-body inverse kinematics (IK) calculation.
The calculated joint angles $\bm{\theta}^{\mathrm{c}}$ are transmitted as commands to the low-level joint position PD controller.
As sensor measurements other than distributed tactile sensors, the control system uses the joint angles from the joint encoders, the contact wrench from the 6-axis force/torque sensors mounted on the feet, and the link orientation from the IMU sensor mounted on the base link.
Tables~\ref{tab:param-mpc-stabilization} and \ref{tab:damping-param} show the control parameters.
We released the control system and the environments of the simulation experiments as open-source code to allow for reproducibility\footnote{\url{https://github.com/mmurooka/MultiContactController/tree/RAL2024}}.

\subsection{Simulation Experiments}

\begin{figure}[tpb]
  \begin{center}
    \includegraphics[width=0.32\columnwidth]{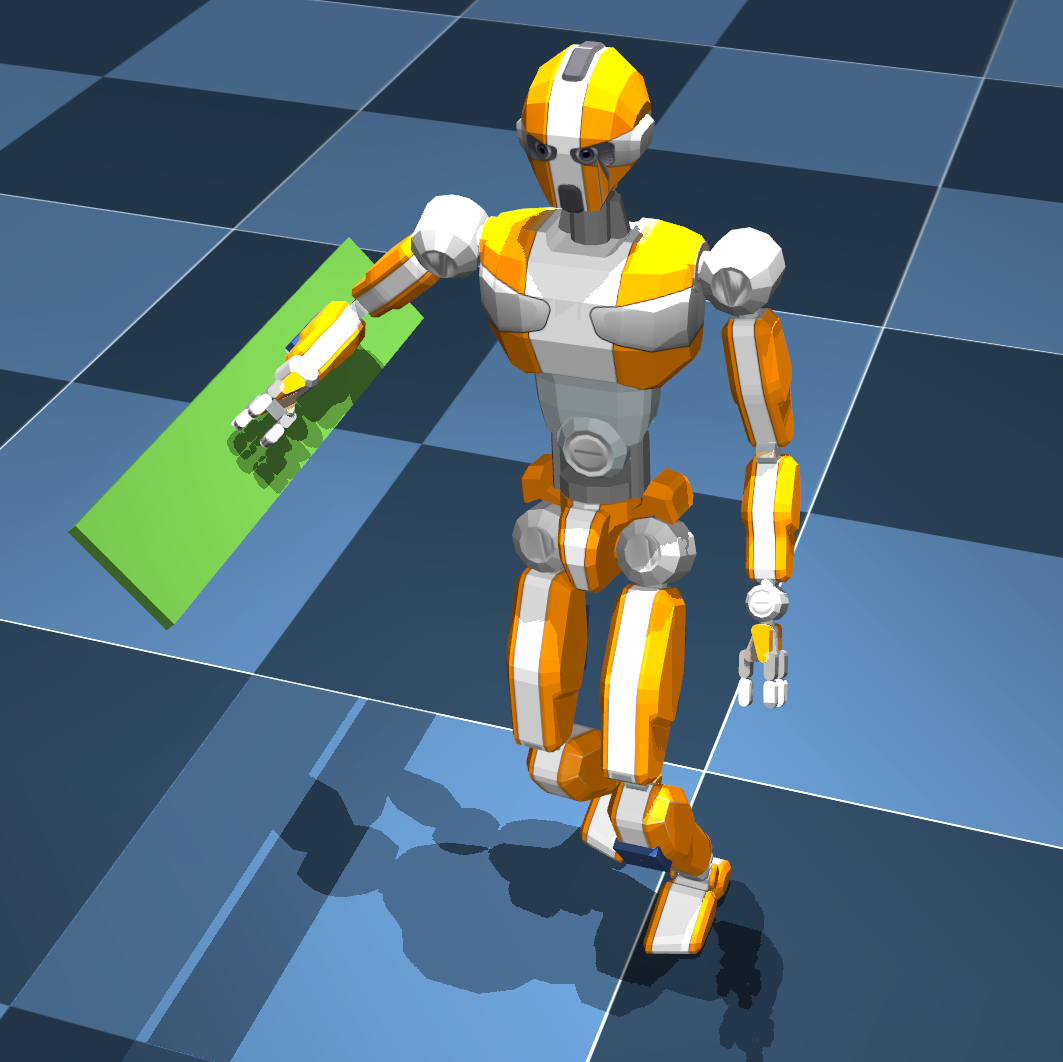}
    \includegraphics[width=0.32\columnwidth]{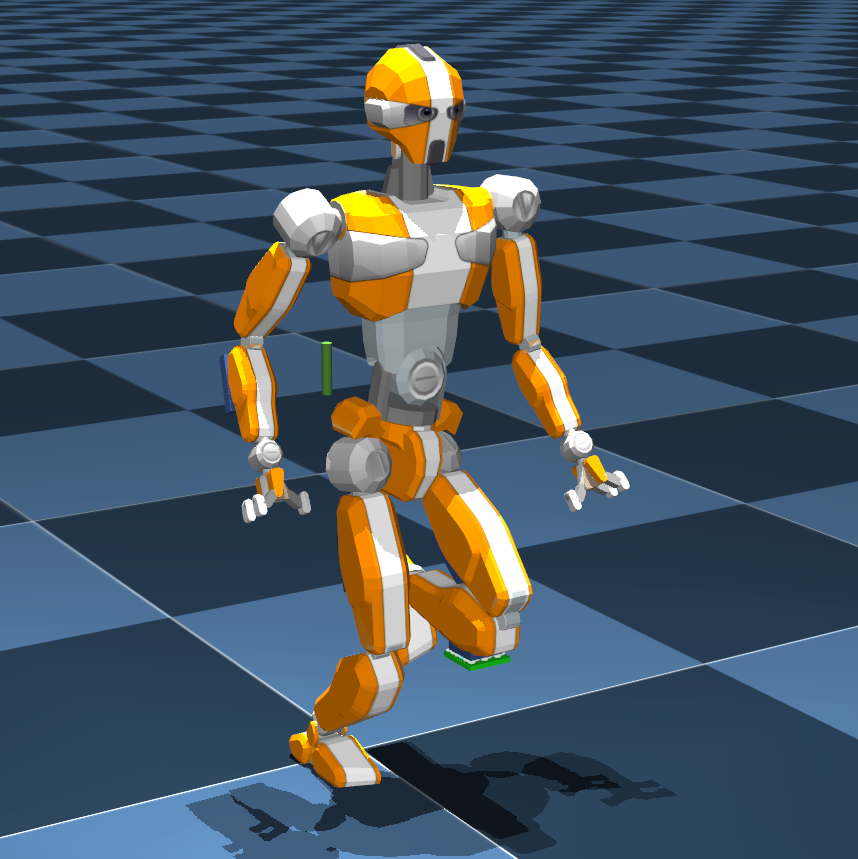}
    \includegraphics[width=0.32\columnwidth]{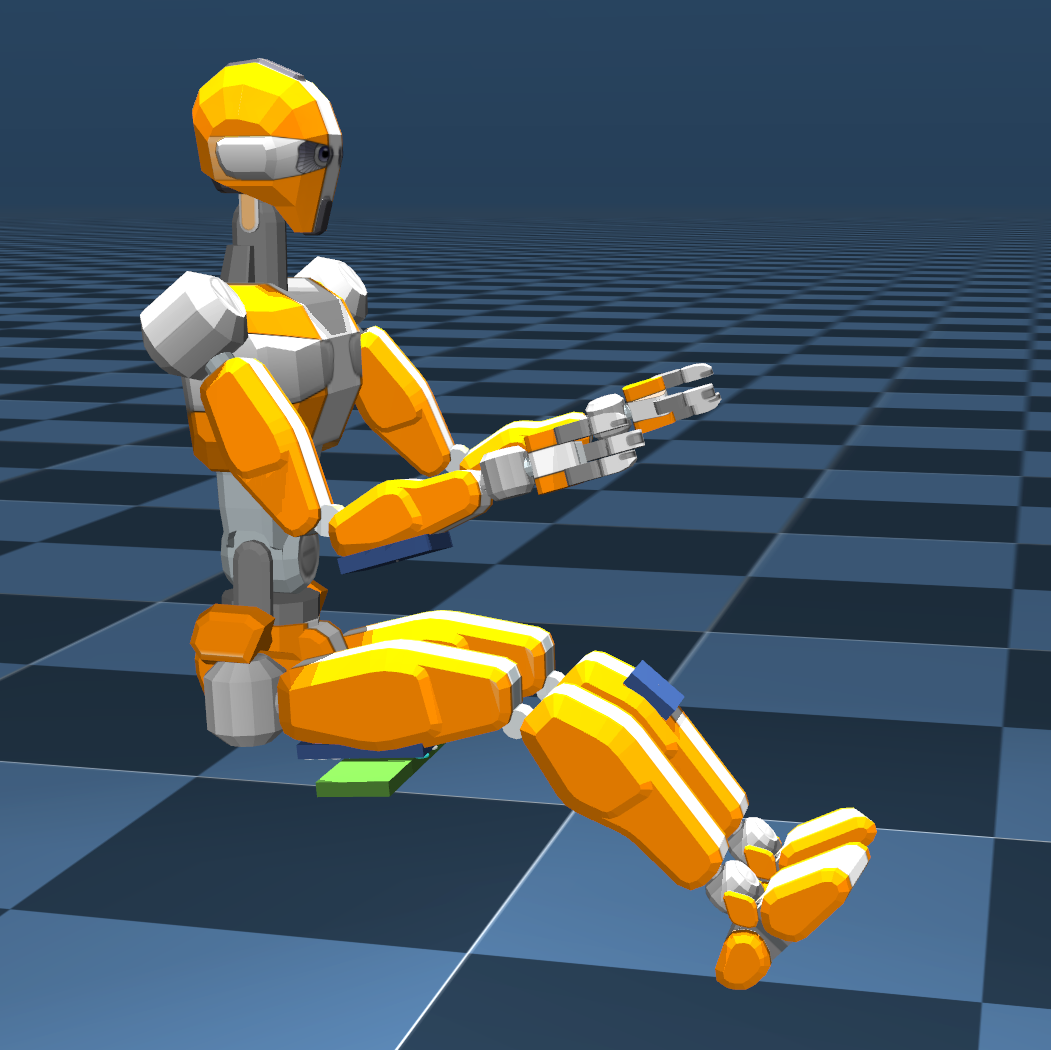}\\
    \vspace{1mm}
    \includegraphics[width=0.32\columnwidth]{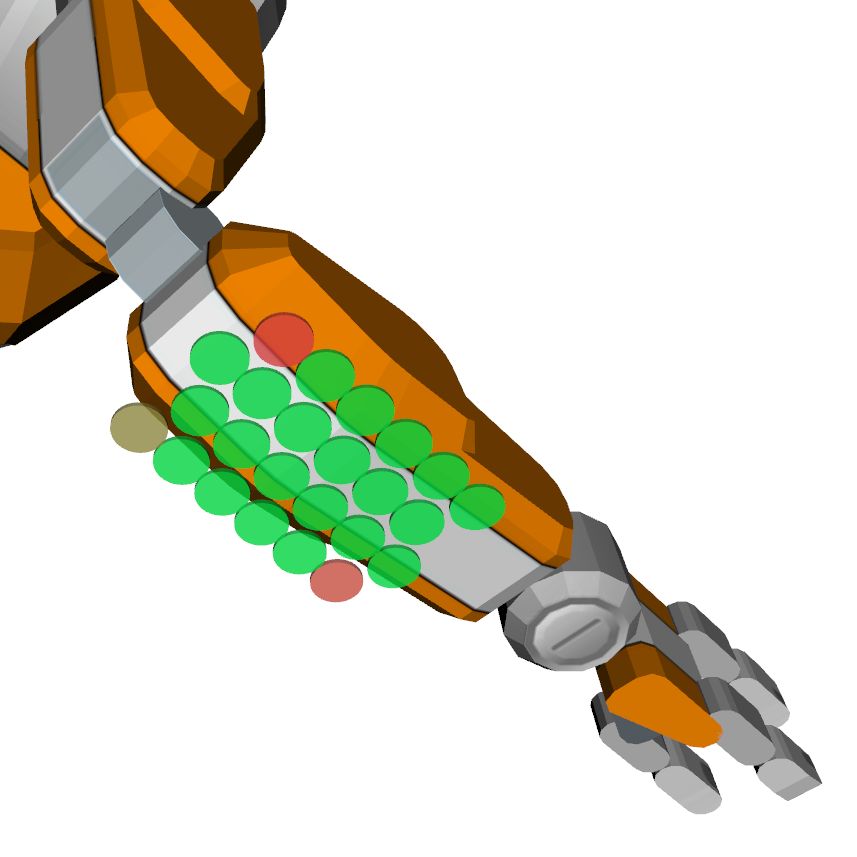}
    \includegraphics[width=0.32\columnwidth]{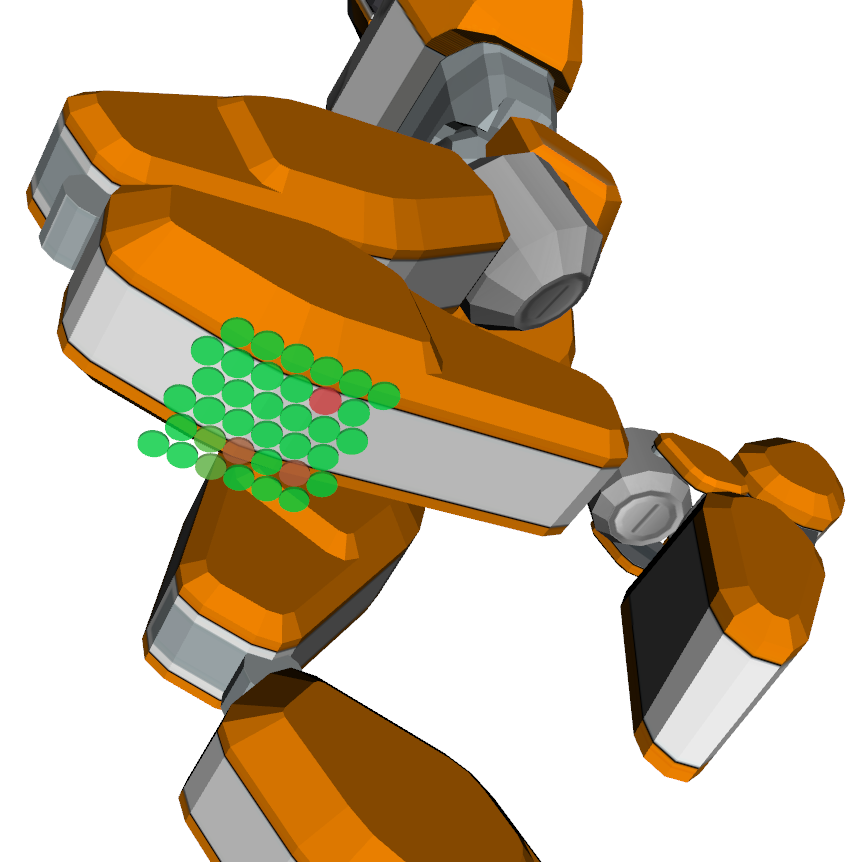}
    \includegraphics[width=0.32\columnwidth]{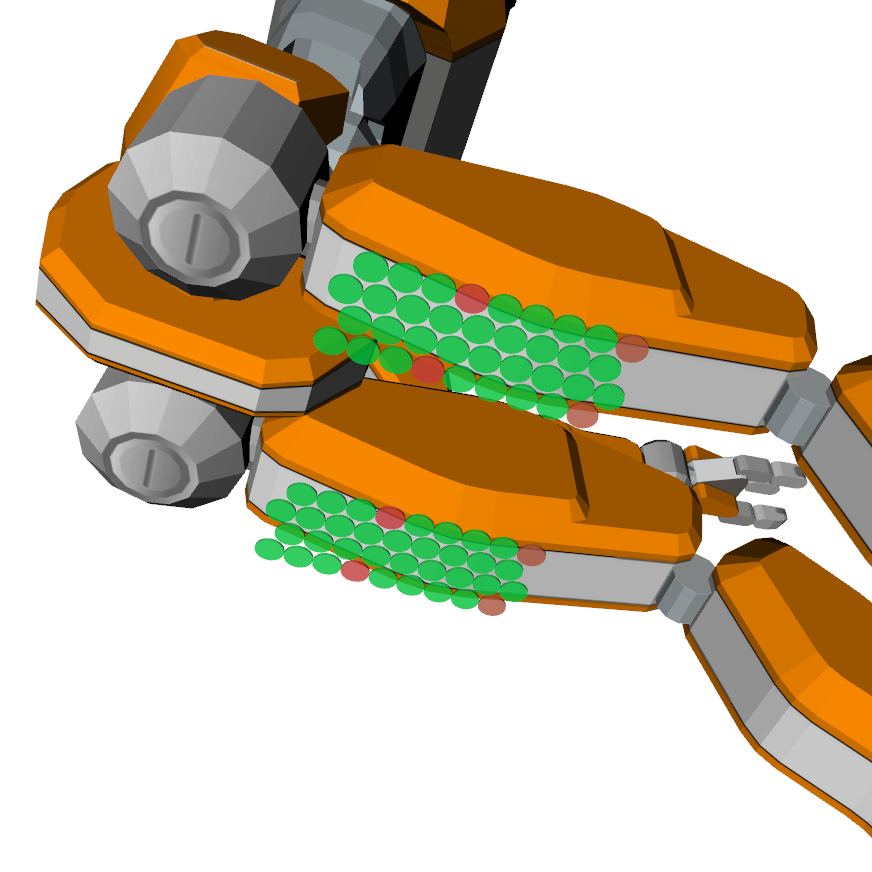}\\
    \vspace{1mm}
    \begin{minipage}{0.32\columnwidth}
      \begin{center} \footnotesize (A) Walking with elbow contact \end{center}
    \end{minipage}
    \begin{minipage}{0.32\columnwidth}
      \begin{center} \footnotesize (B) Standing with knee contact \end{center}
    \end{minipage}
    \begin{minipage}{0.32\columnwidth}
      \begin{center} \footnotesize (C) Sitting with thigh contacts \end{center}
    \end{minipage}\\
    \vspace{1mm}
    \caption{Simulation of whole-body multi-contact motions.
      \newline
      \footnotesize{
        The top row shows snapshots of the simulation, and the bottom row shows the distributed tactile sensors (in each cell, green represents no contact and red represents contact).
        (A) Assuming that there is an obstacle on the left side, the robot walks with the right elbow on the wall, leaning the body to the right.
        (B) The robot stands with the right foot on the ground and the left knee on the block, balancing itself.
        (C) The robot sits on the block with both thighs, keeping its balance.
    }}
    \label{fig:sim-capture}
  \end{center}
\end{figure}
\begin{figure}[tpb]
  \begin{center}
    \includegraphics[width=1.0\columnwidth]{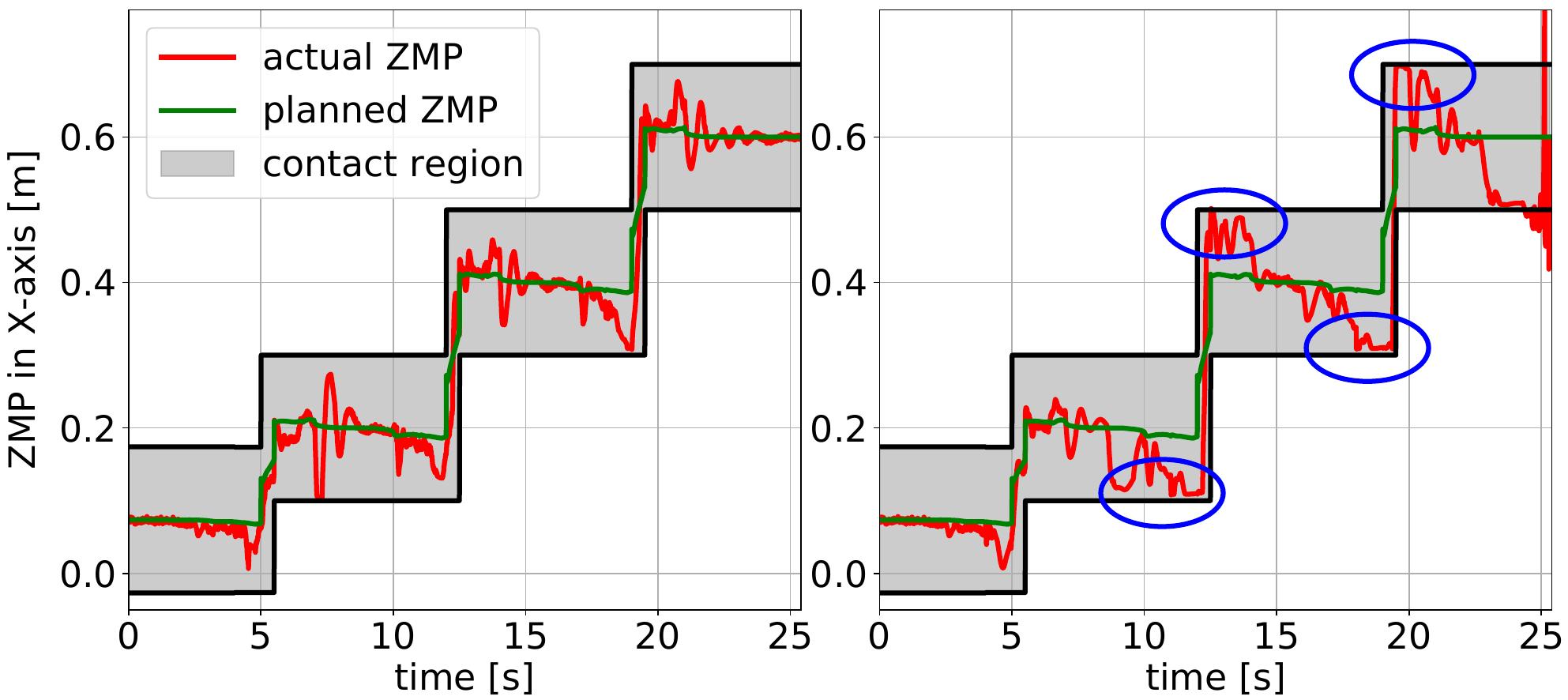}\\
    \begin{minipage}{0.55\columnwidth}
      \begin{center} \footnotesize (A) With tactile feedback \end{center}
    \end{minipage}
    \begin{minipage}{0.42\columnwidth}
      \begin{center} \footnotesize (B) Without tactile feedback \end{center}
    \end{minipage}
    \caption{Measurements when the robot is walking with elbow contact.
      \newline
      \footnotesize{
        The graph shows the feet-only ZMP calculated from the foot contact forces when an error of -0.03~m is added to the inclined wall height in the motion shown in~\figref{fig:sim-capture}~(A).
        In the absence of tactile feedback, the feet-only ZMP reached the edges of the foot contact region (circled in blue in the graph), reducing the robot's stability.
    }}
    \label{fig:sim-elbow-contact}
  \end{center}
\end{figure}
\begin{figure}[tpb]
  \begin{center}
    \includegraphics[width=1.0\columnwidth]{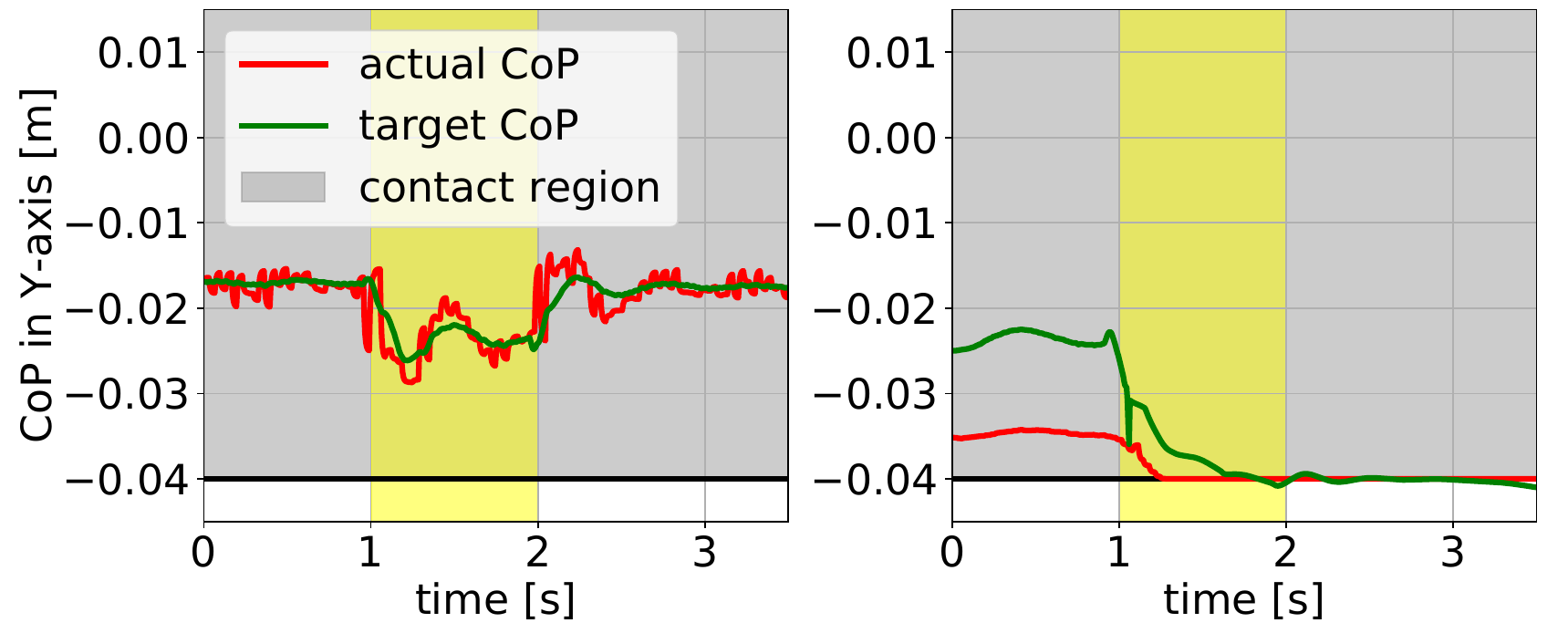}\\
    \begin{minipage}{0.55\columnwidth}
      \begin{center} \footnotesize (A) With tactile feedback \end{center}
    \end{minipage}
    \begin{minipage}{0.42\columnwidth}
      \begin{center} \footnotesize (B) Without tactile feedback \end{center}
    \end{minipage}
    \caption{Measurements when the robot is standing with knee contact.
      \newline
      \footnotesize{
        The graph shows the CoP at the knee contact when a disturbance force of 50~N was applied to the robot during the motion shown in~\figref{fig:sim-capture}~(B).
        The period during which the disturbance force was applied is shaded in yellow.
        In the absence of tactile feedback, the CoP reached the edge of the contact region and the robot fell over.
    }}
    \label{fig:sim-knee-contact}
  \end{center}
\end{figure}
\begin{figure}[tpb]
  \begin{center}
    \includegraphics[width=1.0\columnwidth]{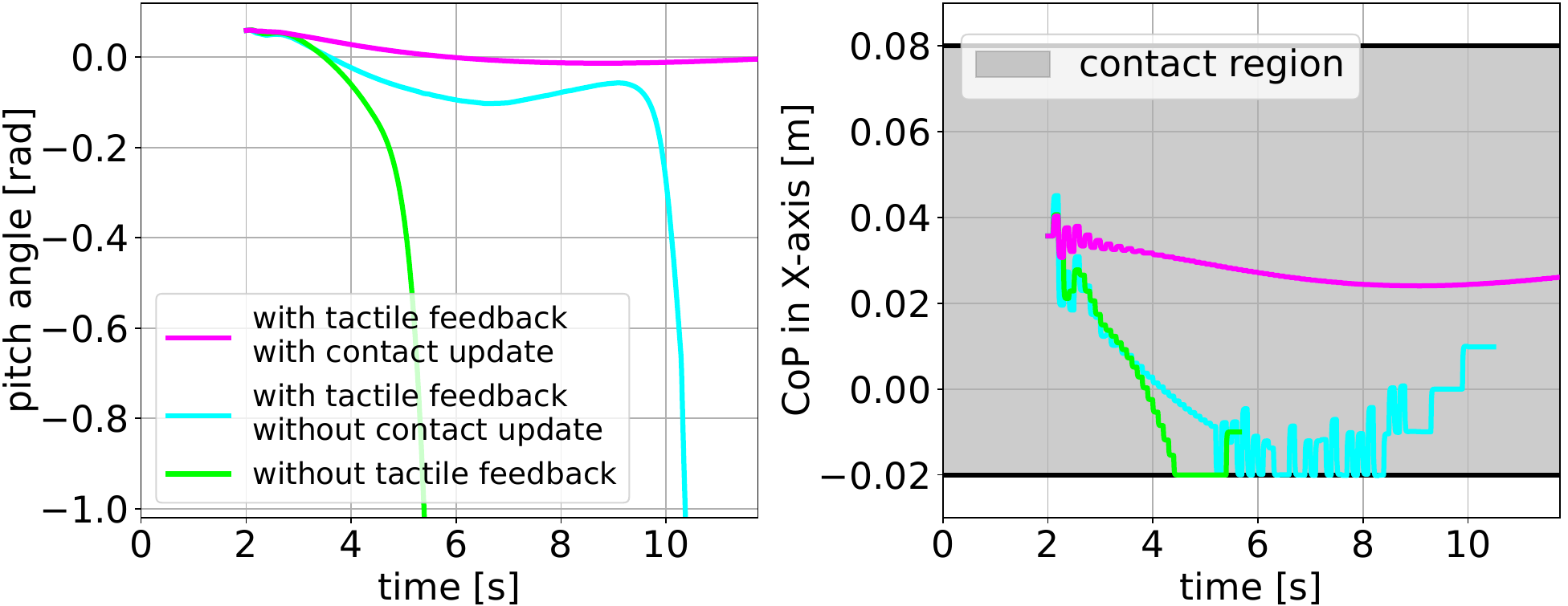}\\
    \begin{minipage}{0.55\columnwidth}
      \begin{center} \footnotesize (A) Pitch angle of the base link \end{center}
    \end{minipage}
    \begin{minipage}{0.42\columnwidth}
      \begin{center} \footnotesize (B) CoP at thigh contacts \end{center}
    \end{minipage}
    \caption{Measurements when the robot is sitting with thigh contacts.
      \newline
      \footnotesize{
        The graph shows the pitch angle of the base link and the CoP at thigh contacts in the motion shown in~\figref{fig:sim-capture}~(C).
        The robot sits on a rotatable seat board, with the dynamics simulation starting at 0~s and the controller starting at 2~s.
        If tactile feedback is not enabled, or if tactile feedback is enabled but the contact region is not updated and a contact region different from the actual one is assumed, the CoP reached the edge of the contact region and the robot fell backward.
    }}
    \label{fig:sim-sitting}
  \end{center}
\end{figure}

We verified various whole-body multi-contact motions by the virtual humanoid robot JVRC1\footnote{JVRC1 is an open-model virtual humanoid robot with a body structure similar to that of HRP-4.} in the dynamics simulator MuJoCo~\cite{MuJoCo:Todorov:IROS2012}.
To simulate distributed tactile sensors mounted on the robot, we implemented a dedicated plugin\footnote{\url{https://github.com/isri-aist/MujocoTactileSensorPlugin}} for MuJoCo.

We validated the effectiveness of the proposed controller with distributed tactile sensors in the three motions shown in~\figref{fig:sim-capture}.
These motions involve contacts at elbows, knees, and thighs, where force/torque sensors are difficult to mount due to geometrical restrictions.
In these experiments, we compared the robustness of motions with and without damping control based on distributed tactile sensors (hereafter referred to as tactile feedback) at these contact areas.

First, in the walking motion with elbow contact shown in~\figref{fig:sim-capture}~(A), we verified whether the robot could walk forward without falling over when an error was added to the height of the inclined wall at elbow level.
The results showed that the error range of the wall height within which the robot could walk was -0.02~m to 0.04~m without tactile feedback, whereas the range was \mbox{-0.03}~m to 0.04~m with tactile feedback.
Furthermore, as shown in \figref{fig:sim-elbow-contact}, tactile feedback improved the tracking performance of the feet-only zero moment point (ZMP)\footnote{ZMP cannot generally be defined under non-coplanar contacts. We define the CoP of both feet calculated from the force/torque sensors mounted on the left and right feet as feet-only ZMP. As long as the feet-only ZMP is within the feet support region, the feet contacts are maintained and the robot does not fall over.}, which contributes to the maintenance of stable feet contacts.
This indicates that tactile feedback enhances robustness against environmental errors.

Second, in the standing motion with knee contact shown in~\figref{fig:sim-capture}~(B), we verified whether the robot could maintain its balance when front-back forces were applied to the base link as disturbances.
The results showed that the robot could withstand disturbance forces from -90~N to 40~N without tactile feedback and from -100~N to 70~N with tactile feedback.
This indicates that tactile feedback enhances robustness against disturbance forces.
\figref{fig:sim-knee-contact} shows the CoP at knee contact.

Third, in the sitting motion with thigh contacts shown in~\figref{fig:sim-capture}~(C), we verified that tactile feedback with contact region update allowed the robot to maintain its balance under the condition that the seat board is movable around the pitch axis with a spring-damper joint.
As shown in~\figref{fig:sim-sitting}, without tactile feedback enabled, a slight shift in the robot's CoM caused the seat board to rotate and the robot fell backward.
In this sitting posture, as shown in~\figref{fig:sim-capture}~(C), only the partial surfaces of the distributed tactile sensors on the robot's thighs were in contact with the seat board.
When the entire surfaces of the distributed tactile sensors were predefined as the contact regions and used as is, the robot fell backward even with tactile feedback enabled.
This was due to the fact that the robot attempted to exert excessive moments at thigh contacts calculated based on the overestimated contact regions, and the contacts could not be maintained.
By updating the contact polygon vertices based on distributed tactile sensor measurements during controller initiation and enabling tactile feedback, the robot successfully maintained its balance on the rotatable seat board.

\subsection{Real-world Experiments}

\begin{figure}[tpb]
  \begin{center}
    \includegraphics[width=0.31\columnwidth]{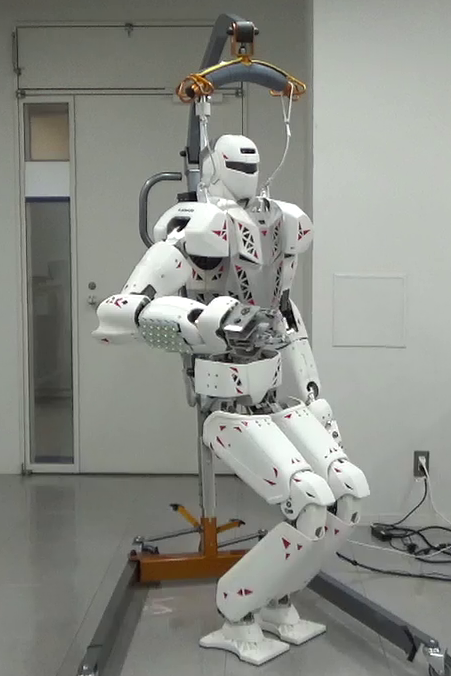}
    \includegraphics[width=0.31\columnwidth]{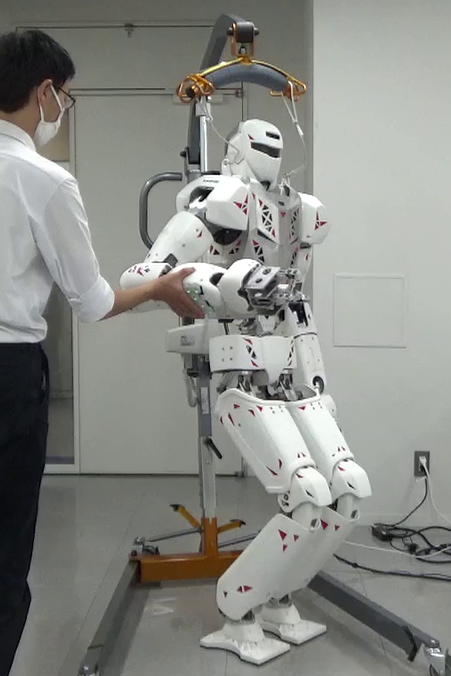}
    \includegraphics[width=0.31\columnwidth]{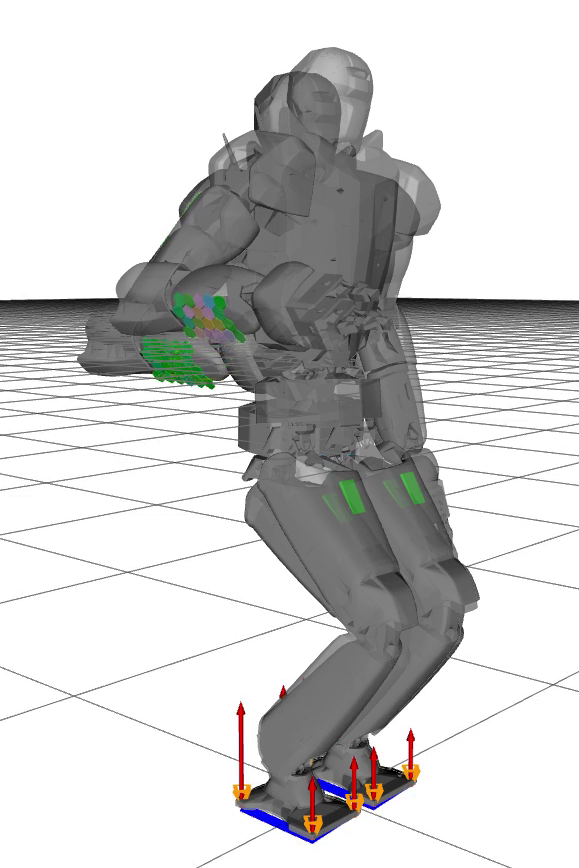}
    \caption{
      Experiment to verify tactile feedback in the forearm.
      \newline
      \footnotesize{
        In response to external force applied to the robot's forearm by the human, the robot updated its whole-body posture to counteract the external force through tactile feedback.
        In the figure on the right, the postures of the robot with and without external force on the forearm are superimposed.
    }}
    \label{fig:impedance-test}
  \end{center}
\end{figure}

We demonstrated whole-body multi-contact motions with a humanoid robot RHP Kaleido equipped with distributed tactile sensors e-skin.
As a preliminary experiment, a human physically interacted with the robot with damping control enabled on the right forearm to verify tactile feedback using distributed tactile sensors.
As shown in~\figref{fig:impedance-test}, the robot's whole body moved in response to the external force applied by the human to follow the target contact force (0~N) on the right forearm, indicating that the limb motion control at an intermediate area, such as the forearm, is working properly.

\begin{figure*}[tpb]
  \begin{center}
    \includegraphics[width=0.49\columnwidth]{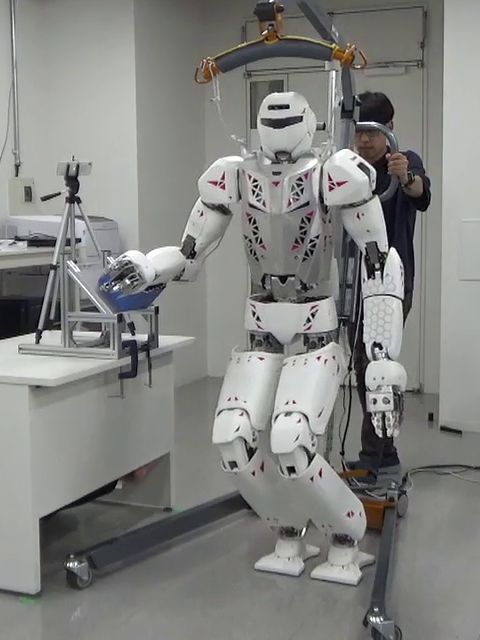}
    \includegraphics[width=0.49\columnwidth]{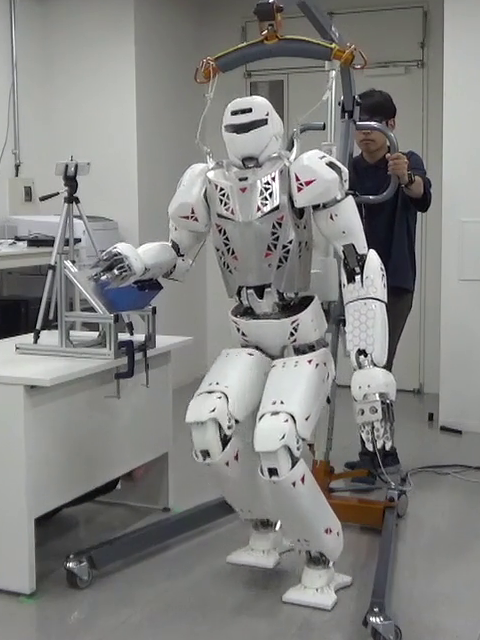}
    \includegraphics[width=0.49\columnwidth]{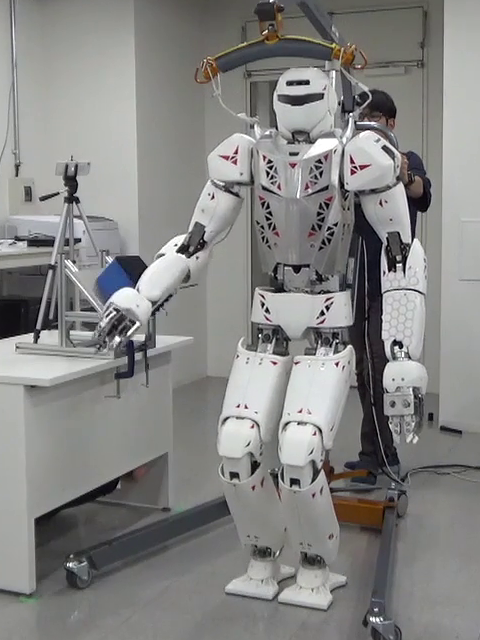}
    \includegraphics[width=0.49\columnwidth]{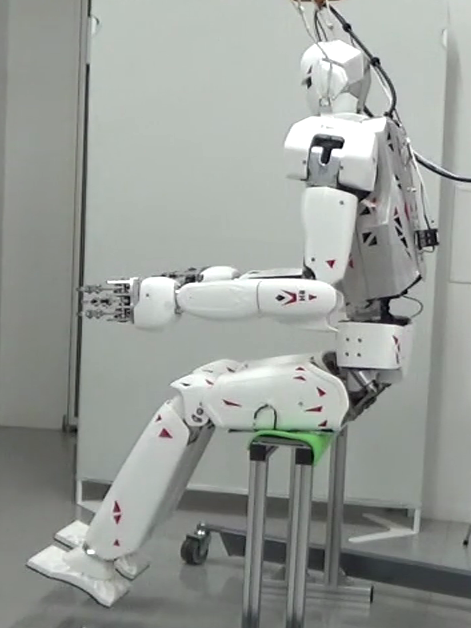}\\
    \begin{minipage}{0.49\columnwidth}
      \begin{center} \footnotesize (A-1) $t = 4$ \lbrack s \!\!\rbrack \end{center}
    \end{minipage}
    \begin{minipage}{0.49\columnwidth}
      \begin{center} \footnotesize (A-2) $t = 7$ \lbrack s \!\!\rbrack \end{center}
    \end{minipage}
    \begin{minipage}{0.49\columnwidth}
      \begin{center} \footnotesize (A-3) $t = 13$ \lbrack s \!\!\rbrack \end{center}
    \end{minipage}
    \begin{minipage}{0.49\columnwidth}
      \begin{center} \footnotesize (B) \end{center}
    \end{minipage}
    \caption{
      Experiments on whole-body multi-contact motions by RHP Kaleido.
      \newline
      \footnotesize{
        (A) RHP Kaleido performed a motion in which the robot contacted the right forearm to the physical environment, leaned to the right side, and took both feet forward one step.
        (B) RHP Kaleido maintains balance in a sitting posture with only both thighs in contact with the seat board.
    }}
    \label{fig:rhp7-elbow-contact}
  \end{center}
\end{figure*}

\begin{figure}[tpb]
  \begin{center}
    \includegraphics[width=1.0\columnwidth]{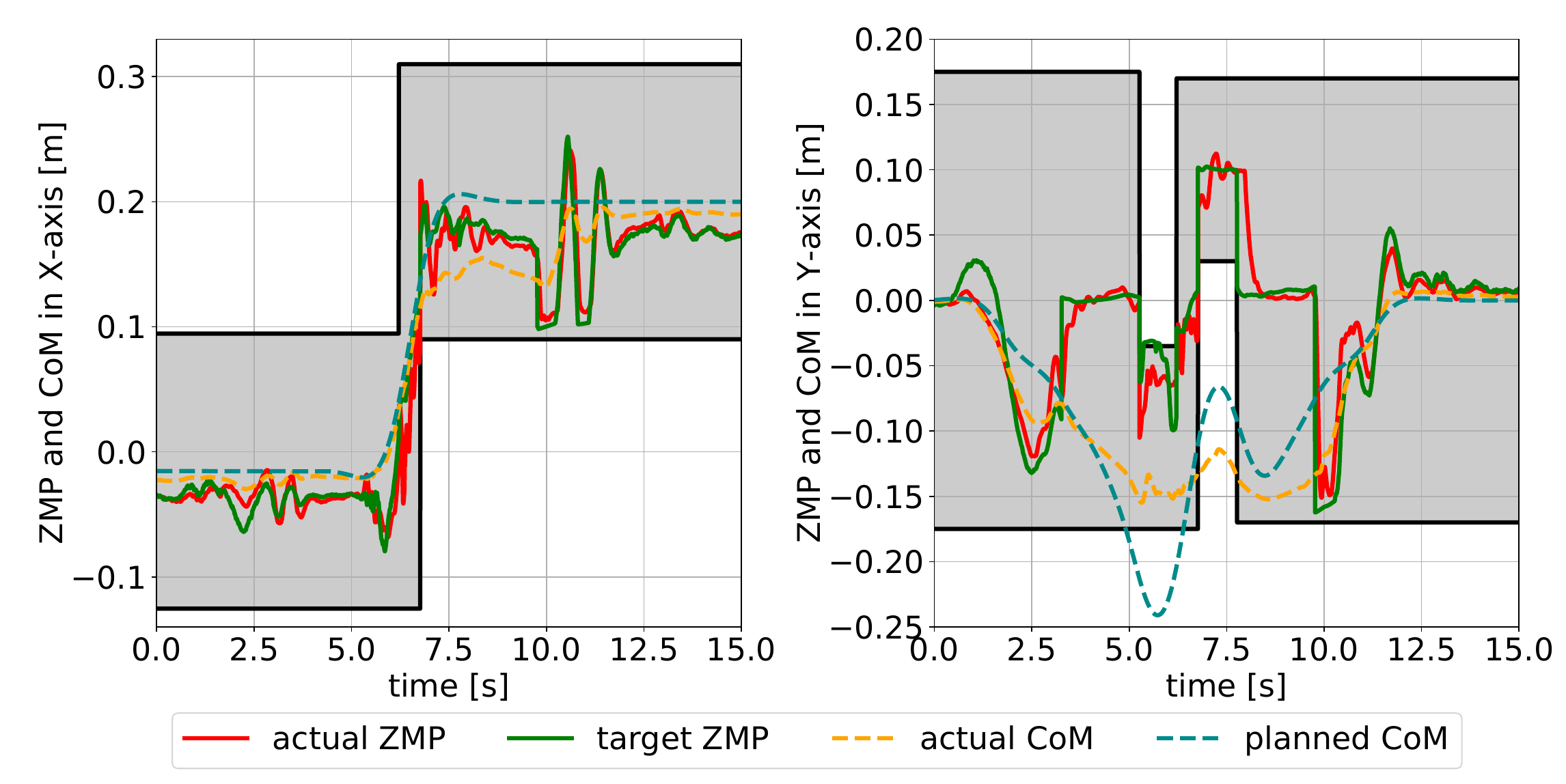}
    \caption{
      ZMP and CoM of the robot during the motion in~\figref{fig:rhp7-elbow-contact}~(A).
      \newline
      \footnotesize{
        The feet-only ZMP in the graph is the CoP of both feet calculated from the force/torque sensors mounted on the left and right feet.
        As long as the feet-only ZMP is within the feet support region (shaded gray in the graph), the feet contacts are maintained and the robot does not fall over.
        Although environmental errors caused deviation between the actual and planned trajectories of the CoM, the robot was still stable from the aspect of the ZMP.
    }}
    \label{fig:rhp7-elbow-contact-zmp}
    \vspace{4mm}
    \includegraphics[width=1.0\columnwidth]{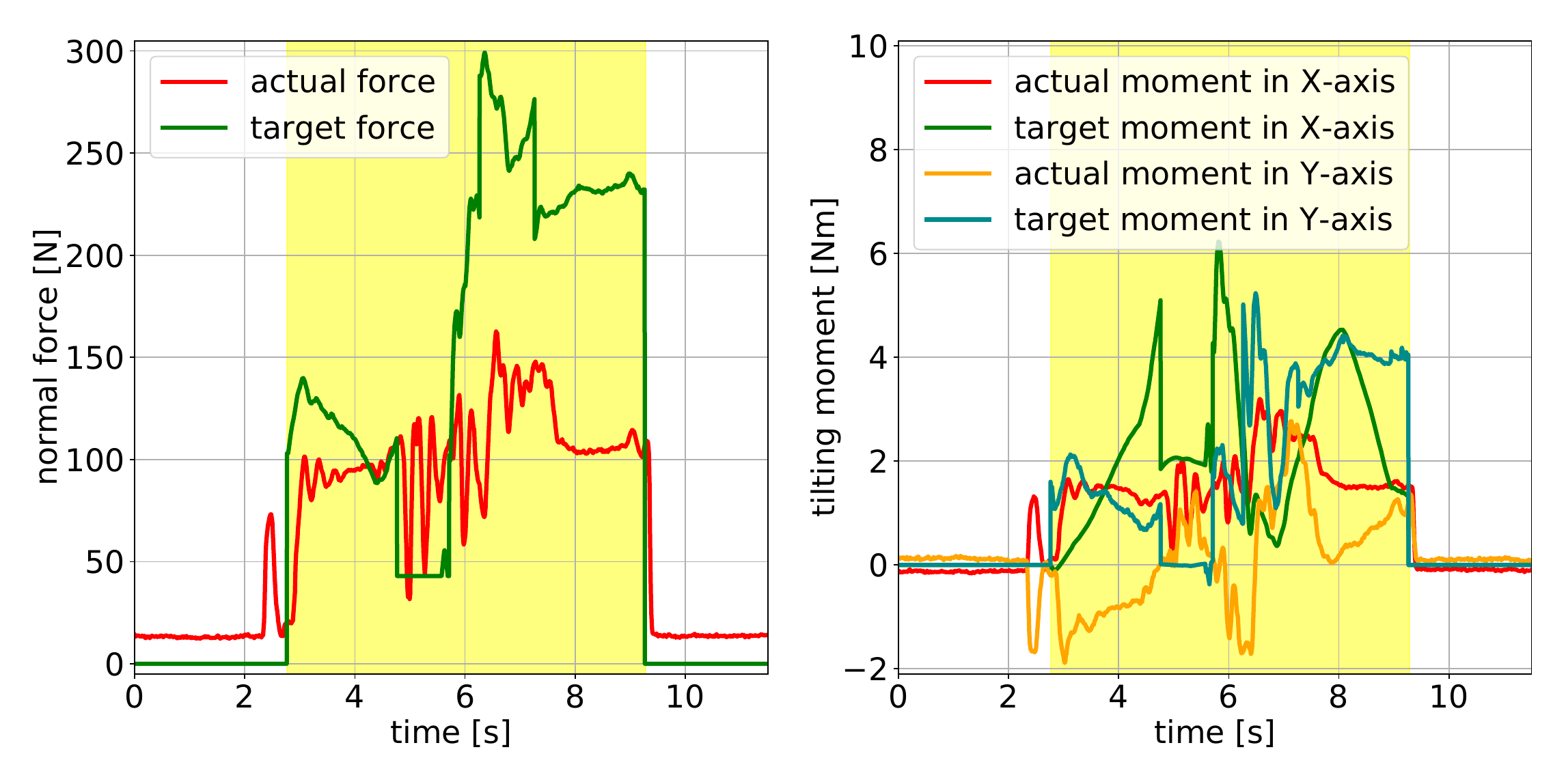}
    \vspace{-6mm}
    \caption{
      Contact wrench of the forearm during the motion in~\figref{fig:rhp7-elbow-contact}~(A).
      \newline
      \footnotesize{
        The graphs represent the force in the normal direction and the moments around the axes parallel to the contact surface of the right forearm.
        Measured values were obtained by distributed tactile sensors mounted on the right forearm.
        The period during which the right forearm was planned to be in contact is shaded in yellow.
    }}
    \label{fig:rhp7-elbow-contact-force}
  \end{center}
\end{figure}


\figref{fig:rhp7-elbow-contact}~(A) shows an experiment of whole-body multi-contact motion, in which the robot steps forward with the forearm in contact with the physical environment (a block mounted to a table).
The robot shifts the CoM about 0.2~m to the right and maintains its balance by exerting a force of about 160~N with the right forearm.
The robot robustly steps forward without being in danger of falling over, because the feet-only ZMP is contained within the feet support region, as shown in~\figref{fig:rhp7-elbow-contact-zmp}.
The contact force and moment of the right forearm have relatively large errors, as shown in~\figref{fig:rhp7-elbow-contact-force}, which are considered to be due to environmental errors (the table on which the block to be contacted was fixed moved slightly due to the impact of the contact) and the nonlinear characteristics of the distributed tactile sensors.
Nevertheless, the generation of contact force begins and ends at the planned timing, and the actual and target contact forces follow approximately the same trend, ensuring stable transition and maintenance of contact. 
This robot motion is a dynamic motion that explicitly considers the effects of centroidal inertia, and cannot be accurately treated with methods from previous studies that assumed a quasi-static motion~\cite{Farnioli:WholebodyLocomanipWalkman:IROS2016,MultiContact:Hiraoka:AR2021}.
We applied preview control~\cite{Motion6DoF:Murooka:RAL2022} instead of DDP as the centroidal MPC described in Section~\ref{subsec:mpc}, taking computational efficiency into consideration to realize real-time control with the robot's on-board computer.

\begin{figure}[tpb]
  \begin{center}
    \includegraphics[width=1.0\columnwidth]{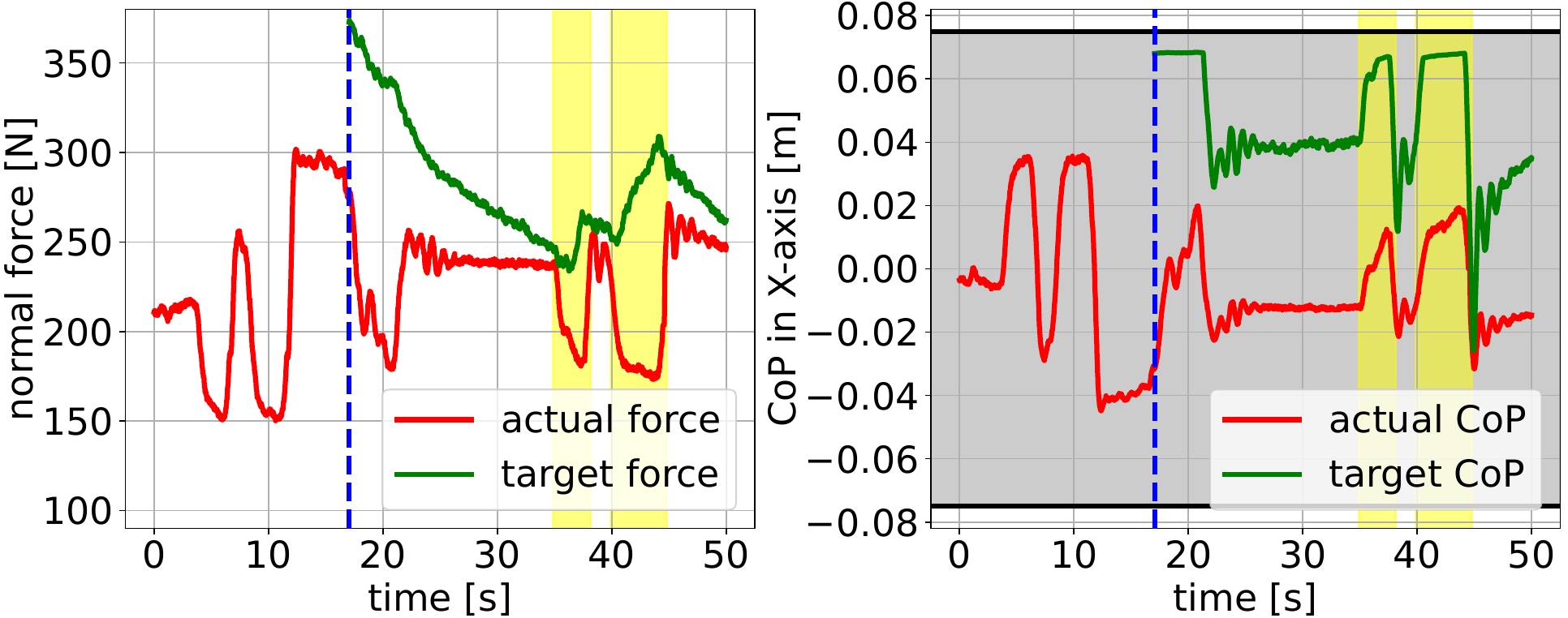}
    \caption{
      Force and CoP of the thighs during the motion in~\figref{fig:rhp7-elbow-contact}~(B).
      \newline
      \footnotesize{
        Before starting to activate the tactile feedback at the timing indicated by the blue line (17~s), the robot was unable to maintain balance by itself and was supported by a human.
        Although errors in the robot model and the environment caused deviations between the actual and target trajectories, by activating tactile feedback, the actual CoP was controlled to stay in the center of the contact region (shaded gray in the graph).
        During the yellow shaded periods, the robot was subjected to a forward disturbance force, but the robot succeeded in maintaining balance.
    }}
    \label{fig:rhp7-sitting-graph}
  \end{center}
\end{figure}

\figref{fig:rhp7-elbow-contact}~(B) shows an experiment in which the robot maintains balance in a sitting posture with only the thighs in contact with the environment.
Without tactile feedback, the robot fell forward due to a slight deviation of the robot's CoM, because the depth of the seat board was as narrow as 15~cm.
When tactile feedback with contact region update was enabled, the robot successfully maintained balance by automatically adjusting the leg joint angles.
The distributed tactile sensors e-skin has tactile and proximity sensors in each cell.
In this experiment, the proximity sensors were used to estimate the contact region because we empirically found that the proximity sensor is more stable than the tactile sensor in estimating the contact region.
\figref{fig:rhp7-sitting-graph} shows the normal force and CoP of the thigh contacts.
The CoP of the thigh contacts is controlled to be at the center of the contact region when the tactile feedback is activated.
The steady error between the actual and target CoP is considered to be due to errors in the robot's mass model and the inclination of the seat surface.

\section{Conclusion}

In this study, we developed a method to realize whole-body multi-contact motion with contacts not only at extremities (e.g., hands and feet) but also at intermediate areas of the limbs (e.g., elbows and knees) on a humanoid robot.
A previous framework of multi-contact motion controller was extended to enable tactile feedback by distributed tactile sensors mounted on the robot's link surfaces.
The controller effectiveness was verified by dynamics simulations to improve the stability of the motions.
Furthermore, we demonstrated that the life-sized humanoid RHP Kaleido, equipped with distributed tactile sensors e-skin, can perform whole-body multi-contact motions such as stepping forward while supporting the body with the forearm and keeping balance in a sitting posture with thigh contacts by the developed method.

This work is an early effort in the important research direction of integrating tactile feedback into real-time balance control in humanoid whole-body contact motion.
To enable the robot's whole-body multi-contact motion to be performed autonomously without prior knowledge of the environment, a planning layer that determines which body parts of the robot should make contact with the environment should be explored in the future.
Since planning contact areas on the robot's whole body causes a combinatorial explosion, it may be useful to consider human motions as a reference.
Another future challenge is the use of more detailed contact information for control, which could contribute to improving the robustness of robot motion.
Although the measurements from distributed tactile sensors were converted to representative values in the form of contact region and wrench in this study, more comprehensive use of the raw measurements may lead to robot motion that adapts more delicately to contact states.
This may require the application of reinforcement or imitation learning based on highly expressive neural networks to overcome the limitations of model-based control methods with respect to optimality and computational cost.

\bibliographystyle{IEEEtran}
\bibliography{main.bib}

\end{document}